\pdfoutput=1

\documentclass[11pt]{article}

\usepackage[hyperref]{EMNLP2022}

\usepackage{times}
\usepackage{latexsym}

\usepackage{times}
\usepackage{latexsym}

\usepackage{enumitem}
\usepackage{multirow}
\usepackage{array}
\usepackage{booktabs} 
\usepackage{microtype}
\usepackage{times}
\usepackage{latexsym}
\usepackage{multirow}
\usepackage{graphics}
\usepackage{booktabs} 
\usepackage{bm}
\usepackage{url}
\usepackage{mathrsfs}
\usepackage{multirow}
\usepackage{balance}
\usepackage{amsthm}
\usepackage{amsmath}
\usepackage{amstext}
\usepackage{amssymb}
\usepackage[ruled,vlined,linesnumbered]{algorithm2e}
\usepackage{subcaption}
\usepackage{overpic}
\usepackage{color}
\usepackage{enumitem}
\usepackage{array}
\usepackage{booktabs} 
\usepackage{enumitem}

\usepackage{microtype}
\usepackage{svg}
\usepackage[T1]{fontenc}
\usepackage[utf8]{inputenc}

\title{Mitigating Shortcuts in Language Models with Soft Label Encoding}

\author{Zirui He\textsuperscript{1}, Huiqi Deng\textsuperscript{2}, Haiyan Zhao\textsuperscript{3}, Ninghao Liu\textsuperscript{4}, Mengnan Du\textsuperscript{3},   \\
  \textsuperscript{1}Stevens Institute of Technology \,
  \textsuperscript{2}Shanghai Jiao Tong University \, \\
  \textsuperscript{3}New Jersey Institute of Technology \,
  \textsuperscript{4} University of Georgia \\
  \small\texttt{zhe22@stevens.edu}, \small\texttt{denghq7@sjtu.edu.cn} \\
  \small\texttt{ ninghao.liu@uga.edu},
  \small\texttt{\{hz54,mengnan.du\}@njit.edu}
}

\date{}

\begin{document}
\maketitle
\begin{abstract} 

Recent research has shown that large language models rely on spurious correlations in the data for natural language understanding (NLU) tasks. 
In this work, we aim to answer the following research question: \emph{Can we reduce spurious correlations by modifying the ground truth labels of the training data?}
Specifically, we propose a simple yet effective debiasing framework, named Soft Label Encoding (SoftLE). We first train a teacher model with hard labels to determine each sample's degree of relying on shortcuts. We then add one dummy class to encode the shortcut degree, which is used to smooth other dimensions in the ground truth label to generate soft labels. This new ground truth label is used to train a more robust student model.
Extensive experiments on two NLU benchmark tasks demonstrate that SoftLE significantly improves out-of-distribution generalization while maintaining satisfactory in-distribution accuracy.
\end{abstract}

\section{Introduction}
Large language models (LLMs), such as BERT~\cite{devlin2018bert}, RoBERTa~\cite{liu2019roberta}, and GPT-3~\cite{brown2020language}, have achieved remarkable performance in various natural language understanding (NLU) tasks. However, recent studies suggest that these LLMs heavily rely on shortcut learning and spurious correlations rather than developing a deeper understanding of language and semantic reasoning across multiple NLU tasks~\cite{niven2019probing,du2021towards,mudrakarta2018did}. This reliance on shortcuts and spurious correlations gives rise to biases within the trained models, which results in their limited generalization capability on out-of-distribution (OOD) datasets.

To mitigate shortcut learning and build more robust models free from biases, several debiasing methods have been proposed, following the framework of knowledge distillation~\cite{hinton2015distilling}. These methods involve training a teacher model with prior knowledge about the task to capture dataset bias and then training a student model to avoid learning the biases present in the teacher model~\cite{he2019unlearn,clark2019don}. However, most existing debiasing methods rely on manual annotation and require specific prior knowledge of bias about the dataset, making it challenging to cover the entire dataset with prior knowledge.

\begin{figure*}[tb]        
\center{\includegraphics[width=16cm]  {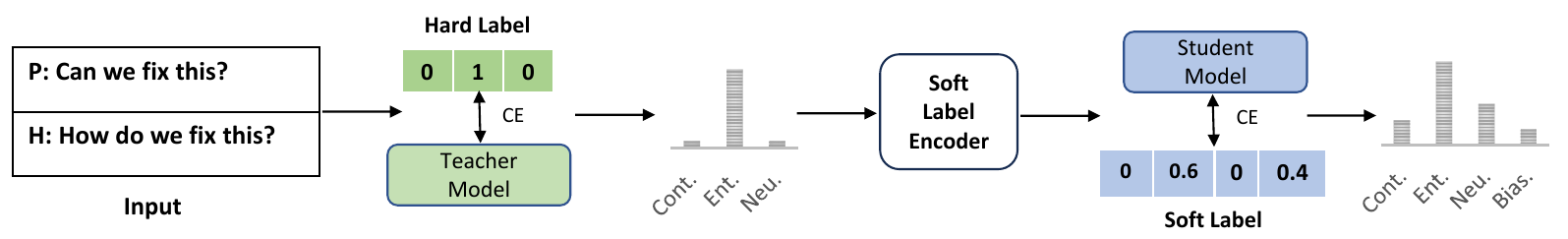}}        
\caption{An overview of the proposed Soft Label Encoding framework.}  
\label{fig:overall-framework}
\end{figure*}

Motivated by the crucial observation that the limited robustness of LLMs on NLU tasks arises from spurious correlations learned during training, we aim to improve generalization and robustness by decreasing the likelihood of learning such correlations through a \underline{data-centric perspective}. One straightforward approach is to filter shortcut features in the input data. However, this method is time-consuming and labor-intensive. This motivates us to explore the following research question: \emph{Can we reduce spurious correlations by modifying the ground truth labels of the training data?}

In this work, we propose a method called Soft Label Encoding (SoftLE) to address the issue of shortcut learning in NLU models through a straightforward \underline{data-centric perspective}. We first train a teacher model with hard labels to determine each sample's degree of relying on shortcuts. We then add one dummy class to encode the shortcut degree, which is used to smooth other dimensions in the ground truth label to generate soft labels. This new ground truth label is used to train a more robust student model. 
The key idea of our method is to reduce spurious correlations between shortcut tokens and certain class labels in the training set. This can be leveraged to discourage models from relying on spurious correlations during model training. This also implicitly encourages the models to derive a deeper understanding of the task. This label smoothing method is efficient since it directly operates on labels and does not require manual feature filtering.
The major contributions of this work can be summarized as:
\begin{itemize}[leftmargin=*]\setlength\itemsep{-0.3em}
\item We propose a data-centric framework, SoftLE, to debias NLU models. 
\item The SoftLE framework is flexible and can be applied to address the shortcut learning issue of various NLU problems.  
\item Experimental results demonstrate that SoftLE improves out-of-distribution generalization while maintaining in-distribution accuracy.
\end{itemize}

\section{Proposed Method}\label{sec2}
In this section, we introduce the proposed Soft Label Encoding (SoftLE) debiasing framework. 

\subsection{SoftLE Debiasing Framework}
\noindent
\textbf{Problem Formulation.}\quad  
NLU tasks are usually formulated as a general multi-class classification problem. Consider a dataset $D = \{(x_i,y_i)\}_{i=1}^N$ consisting of the input data $x_i\in \mathcal{X}$ and the hard labels $y_i\in \mathcal{Y}$, the goal is to train a robust model with good OOD generalization performance. 

\begin{figure}[thb]        
{\includegraphics[width=6.5cm]  {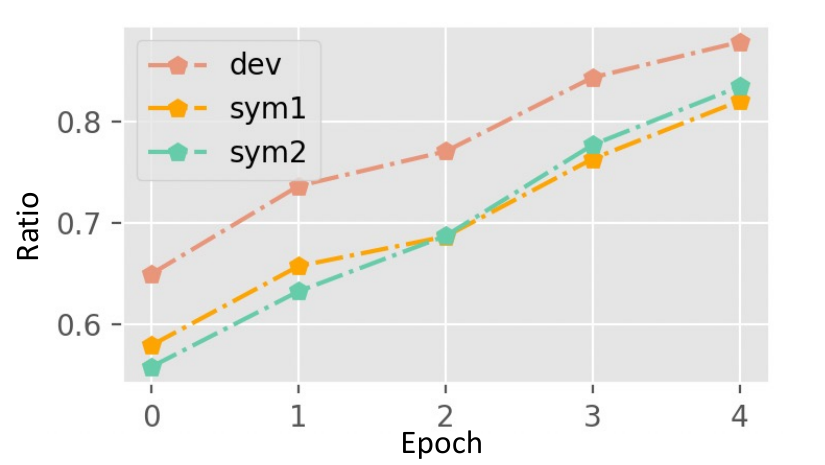}} 
\vspace{-8pt}
\caption{Percentage of over-confident samples in FEVER and corresponding challenge sets during training the teacher model. We set the threshold $\xi$ = 0.9.}
\vspace{-3pt}
\label{pic5}
\end{figure}

\noindent\textbf{Teacher Model Training.}
A biased teacher model $f_T$ containing $K$ classes is first fine-tuned on the corresponding NLU dataset. As shown in Figure~\ref{pic5}, when this model starts to converge, the percentage of over-confident samples in the in-domain set exceeds 0.9, while this ratio is around 0.8 for the OOD sets, indicating there are more over-confident samples in in-domain set.
The in-domain test set contains both shortcut samples and difficult samples, whereas the two OOD sets primarily contain difficult samples. Therefore, the inconsistency in confidence ratios indicates that samples utilizing more shortcut features will be predicted by the teacher model with a higher softmax confidence.
In the following, we leverage the prediction confidence of the model to quantify the degree of shortcut for each training sample.

\noindent\textbf{Quantifying Shortcut Degree.}\,
We fix the parameters of teacher model $f_T$ and calculate the logit value and softmax confidence of training sample $x_i$ as $z_{i}^T$ and $\sigma (z_{i}^T)$. Then, we set the threshold $\xi$ and hyperparameters to calculate the shortcut degree for each over-confident sample(i.e., $\sigma (z_{i}^T)$ > $\xi$):
\begin{equation}
\small
s_{i,j} = \log_{\alpha}(\sigma (z_{i}^T) + \beta).
\label{sl1}
\end{equation}

\noindent\textbf{Soft Label Encoding.}\, Equipped with the shortcut degree, we then transform a K-class classification problem into a K+1 class problem by introducing a new \emph{dummy category}. The value of the dummy category is given as the shortcut degree value $s_{i,j}$. The original label $1$ in one-hot form $y_i$ is transformed into smoothed label: $1-s_{i,j}$.

We illustrate this process using the MNLI task as example (see Figure~\ref{fig:overall-framework}). Here, we set $\xi$ as 0.9. If the teacher model predicts a high softmax confidence $\sigma (z_{i}^T)$ > 0.9 for a sample, then the original three-class classification label $y_i=[0,1,0]$ of this sample will be transformed into a four-class label $y'_i=[0,1-s_{i,j},0,s_{i,j}]$. 

For training samples with a large shortcut degree, more smoothed new labels will be obtained. In contrast, we will preserve the original hard labels for samples with a low shortcut degree.

\subsection{Overall Framework}
We present the overall framework in Algorithm~\ref{alg:SoftLE}. The dummy class is only required during the training stage and will be discarded during inference. 

\noindent\textbf{The Training Stage.}\, We use standard cross entropy loss to train the debiased model:
\begin{equation}
\small
\mathcal{L}_{SL}= -\sum_{i=1}^N \sum_{j=1}^{K+1} y'_{ij} \log (p_{ij}),
\end{equation}
where $p_{ij}$ is the predicted probability for input $x_i$ to have label $j$,
and $y'_{ij}$ is the \emph{transformed label} for training example $i$.

During training, we replace the proposed loss $L_{SL}$ with the standard hard label loss $L_{HL}$ for the first two epochs as a warming-up training: 
\begin{equation}
\small
\mathcal{L}_{HL}= -\sum_{i=1}^N \sum_{j=1}^{K+1} y_{ij} \log (p_{ij}),
\end{equation}
where $y_{ij}$ stands for the \emph{one-hot label} for (K+1)-class classification of the training example.
In the last few epochs, we switch back to using $L_{SL}$. 
This has been demonstrated to retain better ID performance, while achieving similar debiasing performance. We give further analysis in Section~\ref{sec:research-question-1}.

\noindent\textbf{The Inference Stage.}\, We ignore the extra class and predict based on the first $K$ classes:
\begin{equation}
\hat{y}_i = \text{argmax}_{j\in[1, \cdots, K]} p_{i,j}.
\end{equation}

\setlength{\textfloatsep}{14pt}

\begin{algorithm}[t!]\small
\DontPrintSemicolon
\KwIn{Training data $D = \{(x_i,y_i)\}_{i=1}^N$.} Set hyperparameters $\alpha$ and $beta$.\;
 \While {training stage}{
Train teacher network $f_T(x)$. Fix its parameters.
Initialize the student network $f_S(x)$;\;

Shortcut degree: $s_{i,j} = \log_{\alpha}(\sigma (z_{i}^T) + \beta)$;\;
Put $s_{i,j}$ in the dummy class position;\;
Soft label: $1-s_{i,j}$ for the $1$ position of $y_i$;\;
 
Proposed debiasing training loss: $
\mathcal{L}_{SL}= -\sum_{i=1}^N \sum_{j=1}^{K+1} y'_{ij} \log (p_{ij});\;
$
Training loss for first few epochs: $
\mathcal{L}_{HL}= -\sum_{i=1}^N \sum_{j=1}^{K+1} y_{ij} \log (p_{ij}),
$;\;
Specifically, the first two epochs we use $\mathcal{L}_{HL}$, while later epochs we use $\mathcal{L}_{SL}$.
}
\While {inference stage}{
Ignore the dummy class and make the prediction based on values of the first K categories:
$
\hat{y}_i = \text{argmax}_{j\in[1, \cdots, K]} p_{i,j}. 
$
}

\caption{\small Proposed SoftLE framework.}
\label{alg:SoftLE}
\end{algorithm}

\section{Experiments}
In this section, we evaluate the debiasing performance of the proposed SoftLE framework.

\subsection{Tasks and Datasets}
We explore two NLU tasks: natural language inference (NLI) and fact verification. For NLI, we use the MNLI dataset~\cite{williams2017broad} to train biased and de-biased models. We evaluate these models on the in-distribution(ID) MNLI-dev set and the out-of-distribution(OOD) HANS dataset~\cite{mccoy2019right} to test for generalization. 
For fact verification, we use the FEVER dataset~\cite{thorne2018fever} as our ID data. We then evaluate the model's OOD performance on the FEVER symmetric dataset~\cite{schuster2019towards}. Further details are provided in Appendix~\ref{sec:tasks-datasets}.
For both tasks, we employ accuracy as the metric to evaluate the performance of the models on the ID and OOD sets.

\subsection{Comparing Baselines}
We compare our proposed method with four representative baseline methods: \emph{Product of Experts (POE)}~\cite{clark2019don,mahabadi2020end}, \emph{Example Reweighting (ER)}~\cite{schuster2019towards}, \emph{Regularized Confidence}~\cite{utama2020mind}, and \emph{Debiasing Masks}~\cite{Meissner2022DebiasingMA}. More details of the baselines are given in Appendix~\ref{sec:details-baselines}.

\begin{table*}[t]
\centering
\scalebox{0.8}{
\begin{tabular}{lccccccc}
\hline
\multicolumn{1}{c}{}                                             & \multicolumn{3}{c}{MNLI(acc.)}                & \multicolumn{4}{c}{FEVER(acc.)}     \\ \cline{2-8} 
Models                                      & DEV  & HANS & Avg.                      & DEV  & Symm.1 & Symm.2 & Avg. \\ \hline
Original                                                         & \textbf{84.5} & 62.4 & \multicolumn{1}{c|}{73.5} & 85.6 & 55.1   & 62.2   & 67.6 \\
Reweighting~\cite{schuster2019towards}				 & 81.4 & 68.6 & \multicolumn{1}{c|}{75.0} & 84.6 & 61.7   & 64.9   & 70.4 \\
PoE~\cite{clark2019don,mahabadi2020end}    			 & 84.2 & 64.6 & \multicolumn{1}{c|}{74.4} & 82.3 & \textbf{62.0}   & 64.3   & 69.5 \\
Reg-conf~\cite{utama2020mind}             			 & 84.3 & \textbf{69.1} & \multicolumn{1}{c|}{\textbf{76.7}} & 86.4 & 60.5   & 66.2   & 71.0 \\
Debias-Mask~\cite{Meissner2022DebiasingMA} 			 & 81.8 & 68.7 & \multicolumn{1}{c|}{75.3} & 84.6 & -      & 64.9   & -    \\
\textbf{SoftLE}                                                  & 81.2 & 68.1 & \multicolumn{1}{c|}{74.7} & \textbf{87.5} & 60.3   & \textbf{66.9}   & \textbf{71.5} \\ \hline
\end{tabular}}
\caption{\label{t1}Model performance on in-distribution and OOD test set. We select the version that achieves the best performance in the original paper for the listed baseline methods. The Avg. columns report the average score on in-distribution and challenge sets. We highlight the best performance on each dataset.}
\end{table*}

\subsection{Implementation Details}

We employ the Adam optimizer with its default hyperparameters to train the biased model for 5 epochs, where the learning rates for the MNLI and FEVER are set as 5 * $10^{-5}$ and 2 * $10^{-5}$ respectively. In the debiasing model training process, empirical results show that training for 5 epochs with a learning rate of 2 * $10^{-5}$ can provide convergence on FEVER, while 6 epochs are needed to yield the best result on MNLI. Appendix~\ref{sec:details of hyperparameters} is provided for more details. In the following sections, we use BERT-base to test the effectiveness of the proposed debiasing framework (results for RoBERTa-base are given in Appendix~\ref{sec:roberta}).

\begin{figure}[tbp] 
\centering
{\includegraphics[width=6.5cm]  {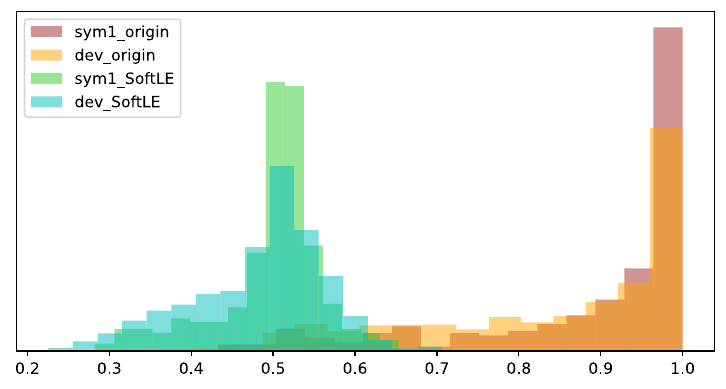}}        
\caption{\label{pic9} We compared the distribution of softmax confidences for the samples misclassified by softLE and the original model (i.e., teacher model) on Fever and Symm.1. Y-axis denotes the ratio.}      
\end{figure}

\subsection{Comparison with Baselines}\label{sec:compare with baselines}
We compare our approach against baselines and report the results in Table~\ref{t1}. We observe that our SoftLE method consistently improves the performance on both challenge sets. However, the in-distribution performance on HANS drops slightly. Figure~\ref{pic9} reveals that despite the biased teacher model assigning high softmax confidences for both in-domain and OOD samples, a larger proportion of high-confidence OOD samples are misclassified. It further illustrates that the SoftLE assigns lower softmax confidences for over-confident samples, thereby effectively reducing the probability of the model incorrectly predicting OOD samples. The results confirm SoftLE leads to a narrower discrepancy between the error rates of ID and OOD sets. 

\subsection{Ablation Study }\label{sec:research-question-1}
In Section~\ref{sec2}, we mentioned that a better trade-off between ID and OOD performance can be achieved by adjusting the loss function during training the debiasing model. To confirm that the combination of this adjustment strategy is necessary to achieve our results, we provide an ablation experiment where the debiasing loss ${L}_{SL}$ is replaced with ${L}_{HL}$ during different training epochs. Previous work has shown that shortcut features tend to be picked up by the NLU model in very early iterations~\cite{du2021towards}. Our results on FEVER support this idea, as shown in Table~\ref{t2}, where we find that replacing training loss in the first 2 epochs outperforms other strategies. As we have observed, SoftLE prevents models from learning spurious correlations, resulting in a lower performance increase on ID and OOD sets during the early stages of training the debiasing model. Thus, this adjustment strategy leverages superficial features, while SoftLE prevents the model from solely relying on superficial features, ultimately achieving a delicate balance.

\begin{table}[tb]
\scalebox{0.8}{
\begin{tabular}{lllll}
\hline
Method                              &  & FEVER & Symm.1 & Symm.2 \\ \hline
Original                       &  & 85.6  & 55.1   & 62.2   \\
SoftLE w/o Replacing &  & 86.6  & 57.7   & 63.9   \\
SoftLE-F2 (Ours)                     &  & \textbf{87.5}  & \textbf{60.3}   & \textbf{66.9}   \\
SoftLE-L2                           &  & 87.1  & 57.8   & 64.4   \\ \hline
\end{tabular}}
\caption{\label{t2}Our experimental results comparing the original method against several loss function adjustment strategies. SoftLE-F2 denotes training with ${L}_{HL}$ for the first 2 epochs, while SoftLE-L2 denotes training with ${L}_{HL}$ for the last 2 epochs.}
\end{table}

\subsection{Why Does Our Algorithm Work? } \label{sec:research-question-2} 
For an over-confident input sample $x = (x^b, x^{-b})$, let $x^b$ denote \textit{biased features} of the sample, and let $x^{-b}$ represent the remaining features of the sample except for the biased ones. It is generally considered that a bias model only uses the biased features $x^b$ to predict the original ground-truth label: 
\begin{equation}
\small
    p(y^{\text{truth}}|x) = 
    p(y^{\text{truth}}|x^b).
\end{equation}
Over-confidence indicates that the predicted probability $p(y^{\text{truth}}|x^b)$ of the sample is very high. 
In other words, for over-confident samples, there is a relatively high spurious correlation between labels and bias features.  

In comparison, when we transform the label of the sample, \textit{i.e.},  altering $y^{\text{truth}}$ to $y^{\text{smooth}}$, it is proved in \cite{clark2019don} that the predictive probability $p(y^{\text{smooth}}|x)$ can be computed as follows:
\begin{equation}
\small
\begin{aligned}
p(y^{\text{smooth}}|x) 
                    & \propto p(y^{\text{smooth}}|x^{-b})p(y^{\text{smooth}}|x^b).
\end{aligned}
\end{equation}
For over-confident samples, we find that the label transformation actually mitigates the potential correlation between labels and biased features. 
In other words, it significantly lowers the predictive probability given biased features, \textit{i.e.}, $p(y^{\text{smooth}}|x^b) < p(y^{\text{truth}}|x^b)$. 
Thus, to maximize $p(y^{\text{smooth}}|x)$, our model has to depend more on unbiased features  $x^{-b}$ to obtain a higher $p(y^{\text{smooth}}|x^{-b})$ value.

\section{Conclusions} 
Recently debiasing NLU tasks has attracted increasing attention from the community. 

In this paper, we proposed SoftLE, a simple yet effective method for mitigating the shortcuts in NLU tasks. We present a theoretical analysis of this approach and experimental results on different NLP benchmark tasks confirming its effectiveness.

\section{Limitations}
There are various ways to quantify the shortcut degree. Our debiasing framework only attempts one solution to generate the shortcut degree of each sample. Going forward, in order to better measure the shortcut degree of the training samples, a more comprehensive analysis is needed. Additionally, although our proposed debiasing framework is general, we have only applied it to two NLU tasks (MNLI and FEVER) and two types of LLMs (i.e., BERT and RoBERTa). In the future, we plan to extend our debiasing framework to more NLU tasks and additional types of LLMs.

\bibliography{acl2020}

\begin{thebibliography}{26}
\expandafter\ifx\csname natexlab\endcsname\relax\def\natexlab#1{#1}\fi

\bibitem[{Brown et~al.(2020)Brown, Mann, Ryder, Subbiah, Kaplan, Dhariwal, Neelakantan, Shyam, Sastry, Askell et~al.}]{brown2020language}
Tom~B Brown, Benjamin Mann, Nick Ryder, Melanie Subbiah, Jared Kaplan, Prafulla Dhariwal, Arvind Neelakantan, Pranav Shyam, Girish Sastry, Amanda Askell, et~al. 2020.
\newblock Language models are few-shot learners.
\newblock \emph{Advances in Neural Information Processing Systems (NeurIPS)}.

\bibitem[{Clark et~al.(2019)Clark, Yatskar, and Zettlemoyer}]{clark2019don}
Christopher Clark, Mark Yatskar, and Luke Zettlemoyer. 2019.
\newblock Don't take the easy way out: Ensemble based methods for avoiding known dataset biases.
\newblock \emph{Empirical Methods in Natural Language Processing (EMNLP)}.

\bibitem[{Devlin et~al.(2019)Devlin, Chang, Lee, and Toutanova}]{devlin2018bert}
Jacob Devlin, Ming-Wei Chang, Kenton Lee, and Kristina Toutanova. 2019.
\newblock Bert: Pre-training of deep bidirectional transformers for language understanding.
\newblock \emph{North American Chapter of the Association for Computational Linguistics (NAACL)}.

\bibitem[{Du et~al.(2022)Du, He, Zou, Tao, and Hu}]{du2022shortcut}
Mengnan Du, Fengxiang He, Na~Zou, Dacheng Tao, and Xia Hu. 2022.
\newblock Shortcut learning of large language models in natural language understanding: A survey.
\newblock \emph{arXiv preprint arXiv:2208.11857}.

\bibitem[{Du et~al.(2021)Du, Manjunatha, Jain, Deshpande, Dernoncourt, Gu, Sun, and Hu}]{du2021towards}
Mengnan Du, Varun Manjunatha, Rajiv Jain, Ruchi Deshpande, Franck Dernoncourt, Jiuxiang Gu, Tong Sun, and Xia Hu. 2021.
\newblock Towards interpreting and mitigating shortcut learning behavior of nlu models.
\newblock \emph{North American Chapter of the Association for Computational Linguistics (NAACL)}.

\bibitem[{He et~al.(2019)He, Zha, and Wang}]{he2019unlearn}
He~He, Sheng Zha, and Haohan Wang. 2019.
\newblock Unlearn dataset bias in natural language inference by fitting the residual.
\newblock \emph{2019 EMNLP workshop}.

\bibitem[{Hinton et~al.(2015)Hinton, Vinyals, and Dean}]{hinton2015distilling}
Geoffrey Hinton, Oriol Vinyals, and Jeff Dean. 2015.
\newblock Distilling the knowledge in a neural network.
\newblock \emph{NeurIPS Deep Learning Workshop}.

\bibitem[{Lapuschkin et~al.(2019)Lapuschkin, Wäldchen, Binder, Montavon, Samek, and Müller}]{Lapuschkin_2019}
Sebastian Lapuschkin, Stephan Wäldchen, Alexander Binder, Gr{\'{e}}goire Montavon, Wojciech Samek, and Klaus-Robert Müller. 2019.
\newblock \href {https://doi.org/10.1038/s41467-019-08987-4} {Unmasking clever hans predictors and assessing what machines really learn}.
\newblock \emph{Nature Communications}, 10(1).

\bibitem[{Liu et~al.(2019)Liu, Ott, Goyal, Du, Joshi, Chen, Levy, Lewis, Zettlemoyer, and Stoyanov}]{liu2019roberta}
Yinhan Liu, Myle Ott, Naman Goyal, Jingfei Du, Mandar Joshi, Danqi Chen, Omer Levy, Mike Lewis, Luke Zettlemoyer, and Veselin Stoyanov. 2019.
\newblock Roberta: A robustly optimized bert pretraining approach.
\newblock \emph{arXiv preprint arXiv:1907.11692}.

\bibitem[{Mahabadi et~al.(2020)Mahabadi, Belinkov, and Henderson}]{mahabadi2020end}
Rabeeh~Karimi Mahabadi, Yonatan Belinkov, and James Henderson. 2020.
\newblock End-to-end bias mitigation by modelling biases in corpora.
\newblock In \emph{58th Annual Meeting of the Association for Computational Linguistics (ACL)}.

\bibitem[{McCoy et~al.(2019)McCoy, Pavlick, and Linzen}]{mccoy2019right}
R~Thomas McCoy, Ellie Pavlick, and Tal Linzen. 2019.
\newblock Right for the wrong reasons: Diagnosing syntactic heuristics in natural language inference.
\newblock \emph{57th Annual Meeting of the Association for Computational Linguistics (ACL)}.

\bibitem[{Meissner et~al.(2022)Meissner, Sugawara, and Aizawa}]{Meissner2022DebiasingMA}
Johannes~Mario Meissner, Saku Sugawara, and Akiko Aizawa. 2022.
\newblock Debiasing masks: A new framework for shortcut mitigation in nlu.
\newblock In \emph{Conference on Empirical Methods in Natural Language Processing}.

\bibitem[{Min et~al.(2020)Min, McCoy, Das, Pitler, and Linzen}]{min-etal-2020-syntactic}
Junghyun Min, R.~Thomas McCoy, Dipanjan Das, Emily Pitler, and Tal Linzen. 2020.
\newblock Syntactic data augmentation increases robustness to inference heuristics.
\newblock In \emph{Proceedings of the 58th Annual Meeting of the Association for Computational Linguistics}.

\bibitem[{Mudrakarta et~al.(2018)Mudrakarta, Taly, Sundararajan, and Dhamdhere}]{mudrakarta2018did}
Pramod~Kaushik Mudrakarta, Ankur Taly, Mukund Sundararajan, and Kedar Dhamdhere. 2018.
\newblock Did the model understand the question?
\newblock \emph{56th Annual Meeting of the Association for Computational Linguistics (ACL)}.

\bibitem[{Niven and Kao(2019)}]{niven2019probing}
Timothy Niven and Hung-Yu Kao. 2019.
\newblock Probing neural network comprehension of natural language arguments.
\newblock \emph{57th Annual Meeting of the Association for Computational Linguistics (ACL)}.

\bibitem[{Sanh et~al.(2021)Sanh, Wolf, Belinkov, and Rush}]{sanh2020learning}
Victor Sanh, Thomas Wolf, Yonatan Belinkov, and Alexander~M Rush. 2021.
\newblock Learning from others' mistakes: Avoiding dataset biases without modeling them.
\newblock \emph{International Conference on Learning Representations (ICLR)}.

\bibitem[{Sanh et~al.(2020)Sanh, Wolf, and Rush}]{sanh2020movement}
Victor Sanh, Thomas Wolf, and Alexander~M Rush. 2020.
\newblock Movement pruning: Adaptive sparsity by fine-tuning.
\newblock \emph{34th Conference on Neural Information Processing Systems (NeurIPS)}.

\bibitem[{Schuster et~al.(2019)Schuster, Shah, Yeo, Filizzola, Santus, and Barzilay}]{schuster2019towards}
Tal Schuster, Darsh~J Shah, Yun Jie~Serene Yeo, Daniel Filizzola, Enrico Santus, and Regina Barzilay. 2019.
\newblock Towards debiasing fact verification models.
\newblock \emph{Empirical Methods in Natural Language Processing (EMNLP)}.

\bibitem[{Thorne et~al.(2018)Thorne, Vlachos, Christodoulopoulos, and Mittal}]{thorne2018fever}
James Thorne, Andreas Vlachos, Christos Christodoulopoulos, and Arpit Mittal. 2018.
\newblock Fever: a large-scale dataset for fact extraction and verification.
\newblock \emph{North American Chapter of the Association for Computational Linguistics (NAACL)}.

\bibitem[{Utama et~al.(2020{\natexlab{a}})Utama, Moosavi, and Gurevych}]{utama2020mind}
Prasetya~Ajie Utama, Nafise~Sadat Moosavi, and Iryna Gurevych. 2020{\natexlab{a}}.
\newblock Mind the trade-off: Debiasing nlu models without degrading the in-distribution performance.
\newblock \emph{58th Annual Meeting of the Association for Computational Linguistics (ACL)}.

\bibitem[{Utama et~al.(2020{\natexlab{b}})Utama, Moosavi, and Gurevych}]{utama2020towards}
Prasetya~Ajie Utama, Nafise~Sadat Moosavi, and Iryna Gurevych. 2020{\natexlab{b}}.
\newblock Towards debiasing nlu models from unknown biases.
\newblock \emph{Empirical Methods in Natural Language Processing (EMNLP)}.

\bibitem[{Williams et~al.(2018)Williams, Nangia, and Bowman}]{williams2017broad}
Adina Williams, Nikita Nangia, and Samuel~R Bowman. 2018.
\newblock A broad-coverage challenge corpus for sentence understanding through inference.
\newblock \emph{North American Chapter of the Association for Computational Linguistics (NAACL)}.

\bibitem[{Xiong et~al.(2021)Xiong, Chen, Pang, Cheng, Ma, and Lan}]{NEURIPS2021_71a8b2ff}
Ruibin Xiong, Yimeng Chen, Liang Pang, Xueqi Cheng, Zhi-Ming Ma, and Yanyan Lan. 2021.
\newblock \href {https://proceedings.neurips.cc/paper_files/paper/2021/file/71a8b2ffe0b594a5c1b3c28090384fd7-Paper.pdf} {Uncertainty calibration for ensemble-based debiasing methods}.
\newblock In \emph{Advances in Neural Information Processing Systems}, volume~34, pages 13657--13669. Curran Associates, Inc.

\bibitem[{Yang et~al.(2019)Yang, Zhang, Tar, and Baldridge}]{pawsx2019emnlp}
Yinfei Yang, Yuan Zhang, Chris Tar, and Jason Baldridge. 2019.
\newblock {PAWS-X: A Cross-lingual Adversarial Dataset for Paraphrase Identification}.
\newblock In \emph{Empirical Methods in Natural Language Processing (EMNLP)}.

\bibitem[{Zellers et~al.(2018)Zellers, Bisk, Schwartz, and Choi}]{zellers2018swag}
Rowan Zellers, Yonatan Bisk, Roy Schwartz, and Yejin Choi. 2018.
\newblock Swag: A large-scale adversarial dataset for grounded commonsense inference.
\newblock \emph{Proceedings of the 2018 Conference on Empirical Methods in Natural Language Processing (EMNLP)}.

\bibitem[{Zhang et~al.(2019)Zhang, Baldridge, and He}]{zhang2019paws}
Yuan Zhang, Jason Baldridge, and Luheng He. 2019.
\newblock Paws: Paraphrase adversaries from word scrambling.
\newblock \emph{arXiv preprint arXiv:1904.01130}.

\end{thebibliography}
\bibliographystyle{acl_natbib}

\clearpage
\appendix

\section{Related Work}
\noindent\textbf{Shortcut Learning Phenomenon.}\quad
Recent studies indicate that shortcut learning has significantly hurt the models’ robustness~\cite{Lapuschkin_2019,niven2019probing,du2022shortcut}.
Fine-tuning pre-trained models can rapidly gain in-distribution improvement while it also gradually increases models' reliance on surface heuristics~\cite{devlin2018bert}. ~\cite{he2019unlearn} have demonstrated that a particular label is highly correlated with the presence of several phrases, independent of the other information provided. Several studies have revealed that artificially constructed samples with heuristic features are very likely to trigger erroneous judgments of the model~\cite{mccoy2019right,schuster2019towards,zhang2019paws,pawsx2019emnlp}.  

\noindent\textbf{Shortcut Learning Mitigation.}\quad
From one perspective, the bias in the dataset can lead to shortcut learning. ~\cite{zellers2018swag} proposed an adversarial filtering method to generate a large-scale data set for the NLI task to reduce annotation artifacts. ~\cite{zhang2019paws,min-etal-2020-syntactic}'s approach aims to remove strongly biased samples from the dataset. However, the criteria for the definition of a bias-free dataset are very obscure. It is also easy to introduce new biases when generating new data.

From another perspective, the scheme of training an automatic de-biasing model is proposed.
Recent studies indicate that inadequate training can be more dependent on simple cues. Training a less capacity model (tiny-BERT) on a full dataset~\cite{sanh2020learning} or a normal model (BERT-base) on a limited dataset~\cite{utama2020towards} has been shown to be effective in leveraging more spurious correlations, which works as an autonomous method to identify the existing biases. A debiasing model is then trained in the framework of knowledge distillation, and the erroneous predictions provided by the biased model will be avoided. An ideal biased model should have learned as many shortcuts as possible. The problem, however, is that it is difficult to determine the degree of bias in the biased model. ~\cite{NEURIPS2021_71a8b2ff} theoretically proved that the debiasing performance can be damaged by inaccurate uncertainty estimations of the bias-only model. ~\cite{utama2020towards,du2021towards} proposed methods that focus on measuring the shortcut degree of each training sample. They also trained bias-only models to identify potentially biased examples and discourage debiasing models from exploiting them throughout the training. More recently, ~\cite{Meissner2022DebiasingMA} introduced a method to avoid retraining a new model by leveraging the movement pruning technique~\cite{sanh2020movement} to remove specific weights of the network that is associated with biased behaviors.

\section{Details of Tasks and Datasets}\label{sec:tasks-datasets}
\textbf{Natural Language Inference (NLI)} is a task that involves understanding and inferring logical relationships between linguistic texts. We select the MNLI dataset~\cite{williams2017broad} to train biased and de-biased models that will evaluate in-distribution performance on the MNLI-dev dataset and out-of-distribution performance on the HANS dataset~\cite{mccoy2019right}. The samples in the HANS dataset are generated by some simple heuristic rules that enable correct classification by relying only on surface features (e.g., word overlap, negation, etc.). The purpose of the HANS dataset is to assess how well the model really performs in terms of inference ability, rather than relying only on shallow surface features.

\noindent\textbf{Fact Verification} is a task that evaluates computer models to make inferences and judgments about the accuracy of a given factual statement. The FEVER dataset~\cite{thorne2018fever} provides claim-evidence pairs and labels for three categories: Supports, Refutes, and NEI(Not Enough Info). The FEVER symmetric dataset~\cite{schuster2019towards} contains 2 subsets with 717 and 712 manually generated claim-evidence pairs, respectively, where the synthetic pairs hold the same relationships (e.g., SUPPORTS or REFUTES) but express different and opposite facts. The goal is to verify whether relying on the cues of the claims leads to incorrect predictions.

\section{Details of Baselines}\label{sec:details-baselines}

\textbf{Product of Experts (POE)}
~\cite{clark2019don,mahabadi2020end} is ensemble learning-based technique where the predictions of multiple "expert" models are combined by taking their product.

\noindent\textbf{Example Reweighting (ER)}
~\cite{schuster2019towards} allocates greater weights to instances of the minority class, consequently incentivizing the model to give increased attention to these instances, thereby enhancing its capacity to accurately identify the less represented class.

\noindent\textbf{Regularized Confidence}
~\cite{utama2020mind} is motivated by the fact that overconfidence can indicate that the model is not well-calibrated and may perform poorly on unseen data. To address this issue, a regularization term is added to the loss function to encourage the model to output a more uniform (or less confident) probability distribution.

\noindent\textbf{Debiasing Masks}
~\cite{Meissner2022DebiasingMA} removes specific weights of the network that is associated with biased behavior without altering the original model. A mask search is performed to identify and remove those weights.

\section{More Experimental Details}\label{sec:details of hyperparameters}
We consider the following values to set the over-confident threshold $\xi$: 0.86, 0.88, 0.90, 0.92, and 0.94. Figure~\ref{pic5} shows the percentage of over-confident samples predicted by the biased model on the FEVER and its challenge sets for a threshold of 0.9. We conducted several experiments to verify that changing this threshold only slightly changed the proportion of over-confident samples. For the FEVER and MNLI datasets, we set the threshold values to 0.88 and 0.9 respectively to obtain the best debiasing effect.

The two hyperparameters $\alpha$ and $\beta$ in Equation~\ref{sl1} are designed to map the extremely high softmax confidences(i.e., from 0.9 to 1) to a larger range. Considering that our goal is to smooth the softmax confidence to about 0.6, we set $\alpha$ = 1.48 and $\beta$ = 0.2 to yield better results. (i.e., for an over-confident sample with softmax confidence of 0.99, its shortcut degree will be calculated as $\log_{1.48}( 0.99 + 0.2 )$ = 0.444. )

The implementation is based on Python and utilizes the Hugging Face package. The embedding size was set to 16. For the debiasing model, the learning rate was linearly increased for 2000 warming steps and then gradually decreased to 0. Conversely, the biased teacher model did not require any warming steps. The training process on the A100 GPU took approximately seven hours, primarily due to the larger training data in MNLI, whereas training on FEVER was completed in four hours. It is worth noting that this time refers to the total duration for training both the teacher model and the debiasing model. Once you obtain a suitable teacher model, you can utilize it to fine-tune various debiasing models, which will reduce the time required by half.

\section{Results over RoBERTa-base}\label{sec:roberta}
Table~\ref{t1} presents the results of the proposed debiasing framework using the BERT-base model. We also test the framework using RoBERTa-base~\cite{liu2019roberta}. The results are provided in Table~\ref{t3} for the FEVER task. The results indicate that our proposed method could improve generalization over two challenging OOD test sets while having only a minor sacrifice on the in-domain test set.

\begin{table}[thb]
\scalebox{0.8}{
\begin{tabular}{lllll}
\hline
Method           &  & FEVER & Symm.1 & Symm.2 \\ \hline
Original-RoBERTa &  & 88.1  & 59.2   & 64.7   \\
SoftLE-RoBERTa   &  & 87.9  & 63.2   & 67.5   \\ \hline
\end{tabular}}
\caption{\label{t3}We validated the effectiveness of SoftLE using RoBERTa-base model on FEVER dataset.}
\end{table}

\begin{figure}[tb]        
\center{\includegraphics[width=7.5cm]  {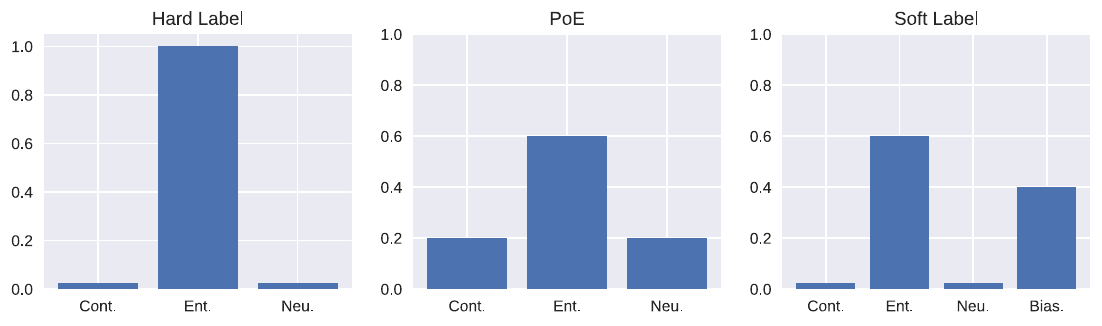}}        
\caption{\label{pic1} An intuitive comparison between SoftLE framework with original method and Product of Experts (POE). Y-axis denotes softmax confidence.}    
\end{figure}

\section{Difference between SoftLE and POE?} \label{sec:research-question-3}
Recent study indicates that in the PoE training, the debiasing model is encouraged to learn from the errors of the biased model instead of mimicking~\cite{sanh2020learning}. However, Figure~\ref{pic9} reveals that there exists a gap between misclassified samples of ID and OOD sets. The abuse of spurious correlations results in more misclassified samples with high softmax confidence in the OOD set, and is not consistent with the cause of errors in the ID set. A key difference thus arises: SoftLE method scale down all the over-confident samples while these samples are less focused in PoE training. We also provide an intuitive comparison between the proposed SoftLE framework with standard model training and PoE in Figure~\ref{pic1}. The results demonstrate that SoftLE training attains a comparable outcome to the PoE approach concerning confidence reduction. However, the introduction of bias scores reveals its potential to enhance the model's ability to autonomously acquire biases in datasets and focus on the intended NLU tasks.

\end{document}